\documentclass[runningheads]{llncs}

\usepackage[T1]{fontenc}
\usepackage{graphicx}
\usepackage{booktabs}
\usepackage{amsmath,amssymb}
\usepackage{multirow}
\usepackage{url}
\usepackage{tabularx}
\usepackage[pagebackref=true,breaklinks=true,colorlinks,bookmarks=false]{hyperref}
\usepackage{orcidlink}
\usepackage{cleveref}

\usepackage{amsmath}
\usepackage{enumitem}
\usepackage{microtype}

\makeatletter
\g@addto@macro\normalsize{%
  \setlength\abovedisplayskip{2pt plus 1pt minus 1pt}%
  \setlength\belowdisplayskip{2pt plus 1pt minus 1pt}%
  \setlength\abovedisplayshortskip{0pt plus 1pt}%
  \setlength\belowdisplayshortskip{1pt plus 1pt minus 1pt}%
}
\g@addto@macro\small{%
  \setlength\abovedisplayskip{2pt plus 1pt minus 1pt}%
  \setlength\belowdisplayskip{2pt plus 1pt minus 1pt}%
  \setlength\abovedisplayshortskip{0pt plus 1pt}%
  \setlength\belowdisplayshortskip{1pt plus 1pt minus 1pt}%
}
\makeatother

\setlist{nosep, topsep=2pt, partopsep=0pt, itemsep=1pt,
         parsep=0pt, leftmargin=*}

\newcommand{\myorcid}[1]{\raisebox{0.3ex}{\orcidlink{#1}}}
\newcommand{\entity}{\mathcal{E}}

\begin{document}

\title{Entity-Constrained CBCT Retrieval for Low-Resource Dental Record Completion}
\titlerunning{Entity-Constrained CBCT Retrieval}

\author{
Nhi Ngoc-Yen Nguyen\inst{1,*}
\myorcid{0009-0009-3640-8597}
\and
Thai Nguyen\inst{1,2,*}
\myorcid{0009-0001-8546-8838}
\and
Kiet Huynh\inst{1}
\myorcid{0009-0003-5665-3406}
\and
Huy-Hieu Pham\inst{1,\dagger}
\myorcid{0000-0003-4851-2518}
}

\authorrunning{Nguyen et al.}

\institute{
VinUni-Illinois Smart Health Center, VinUniversity, Hanoi, Vietnam
\and
Vietnam National University Ho Chi Minh City, University of Science, Vietnam\\
\email{\{nhi.nny, thai.nt, hieu.ph\}@vinuni.edu.vn}\\
\email{hctkiet22@clc.fitus.edu.vn}
}

\maketitle
\renewcommand\thefootnote{}
\footnotetext{* Equal contribution.}
\footnotetext{\textdagger\ Corresponding author: hieuph@vinuni.edu.vn}

% =====================================================================
\begin{abstract}
Completing dental records from cone-beam computed tomography (CBCT) is difficult when annotation is scarce and individual clinical fields are supported by different types of evidence. MMDental Task~3 requires seven-field record completion from only 50 labeled CBCT cases and scores the correctness of structured FDI positions and ICD codes; consequently, a visually plausible retrieved record can still be harmful when it introduces an unsupported entity. We propose \emph{Entity-Constrained CBCT-Guided Retrieval} (ECCR), a parameter-free framework that separates evidence availability from evidence authority. A corpus-derived prior first supplies the complete record. A frozen 3D encoder retrieves image-conditioned Diagnosis evidence, which is appended only if it does not expand the prior FDI or ICD entity set, so the asserted entity set is invariant by construction. On public validation, ECCR reaches a weighted score of $0.3134$, improving on both full-record multimodal retrieval ($0.2237$) and a static text-only prior ($0.2915$); the guard blocks $63.3\%$ of retrieved candidates, each of which would otherwise have injected an FDI position or ICD code absent from the prior. On the final test evaluation, ECCR obtains $11.37$ of a $97.4$-point attainable maximum, securing second place overall. The result indicates that, in an extreme low-resource setting, controlling what multimodal evidence is allowed to modify can be more reliable than transferring an entire retrieved record.
\keywords{Multimodal generation \and Evidence-constrained retrieval \and Cone-beam CT \and Clinical record completion \and Low-resource learning}
\end{abstract}

% =====================================================================
\section{Introduction}
\label{sec:introduction}

Cone-beam computed tomography (CBCT) is essential for dental diagnosis, yet converting 3D volumes into structured 7-field clinical records is challenging---particularly under the low-resource regime of MMDental Task~3 (50 labeled cases). Granting retrieved evidence equal authority across fields risks hallucinating findings or overwriting accurate text, directly incurring penalties on strictly evaluated FDI tooth positions and ICD codes.

While medical vision--language models enable image-conditioned retrieval \cite{bai2024m3d,xin2025med3dvlm,hamamci2024ct2rep,hosseini2025samf}, unrestricted transfer fails under sparse supervision. In public validation, replacing records with M3D-CLIP neighbors yields 0.2237, underperforming a text-only BLEU-medoid prior (0.2915). Unrestricted retrieval can therefore provide useful context without being reliable enough to control every clinical field.

We propose \emph{Entity-Constrained CBCT-Guided Retrieval} (ECCR). ECCR establishes a corpus-derived prior for all seven fields, retrieves a CBCT candidate via a frozen 3D encoder, and strictly limits updates to the Diagnosis text---accepting them only if no new FDI positions or ICD codes are introduced. Our contributions are:
\begin{enumerate}
\item We formulate low-resource CBCT record completion as an evidence-authority problem, separating retrieval capability from field modification permissions.
\item We introduce a parameter-free entity-constrained admission rule, with a proof that the asserted entity set is invariant under the update.
\item We provide challenge-oriented analysis through authority comparisons, update traces, offline ablations, and a full sub-metric decomposition of our second-ranked result, including a frank account of where the design loses points.
\end{enumerate}

% =====================================================================
\section{Related Work}
\label{sec:related}

\paragraph{3D medical image--text modelling.}
Large 3D vision--language models such as M3D and Med3DVLM connect volumetric images with medical text \cite{bai2024m3d,xin2025med3dvlm}, and report generation frameworks such as CT2Rep and SAMF retrieve or synthesize radiological reports \cite{hamamci2024ct2rep,hosseini2025samf}. These models make image-conditioned evidence accessible but do not delineate which fields can be safely modified under extreme data scarcity. We leverage frozen 3D representations while explicitly governing field-level updates.

\paragraph{Retrieval, factuality, and structured evidence.}
Retrieval-augmented generation improves access to relevant evidence \cite{lewis2020rag} but risks introducing unsupported clinical findings. Prior work enforces factuality through specialized training objectives \cite{miura2021factual,delbrouck2022semantic} or structured representations such as RadGraph and knowledge graphs \cite{jain2021radgraph,zhang2020kgreport}. Since training a dedicated factuality model is infeasible with 50 cases, ECCR instead uses extracted FDI positions and ICD codes as a parameter-free, deterministic admission rule.

% =====================================================================
\section{Dataset Analysis and Design Motivation}
\label{sec:data}

The challenge data contain labeled training, unlabeled training, and public-validation splits, with one NIfTI CBCT volume per case. The labeled CSV comprises 161 visit rows from 50 cases; we aggregate rows sharing a filename by concatenating their unique, non-empty field values, forming one structured record per volume.

\paragraph{Field heterogeneity.}
\Cref{tab:fields} shows why a uniform update policy is unsuitable. Doctor advices is strongly templated (59 unique values, one phrase appearing 33 times), Handle and Oral check are almost visit-specific, and Main appeal and Present medical history are frequently absent. Image evidence should therefore not be granted the same authority across all fields.

\paragraph{Structured-entity skew.}
The 759 tooth mentions cover 45 FDI positions and the 116 ICD mentions cover 29 codes, both concentrated: teeth 16, 46, 17, 26, 36, 21, 38 and codes K00.002, K05.300, K07.302 dominate. This concentration makes a corpus-derived text prior useful, while the structured scoring makes unsupported entity transfer costly.

\paragraph{Geometry.}
All 50 labeled volumes share geometry $640\times640\times400$ at isotropic $0.25$~mm spacing, so geometry does not explain variation in this split. Global resizing and pooling can obscure tooth-localized findings, motivating retrieval as optional evidence rather than complete-record replacement.

\begin{table}[t]
\caption{Field heterogeneity motivates entity-constrained retrieval update. Statistics cover 161 labeled visits; Filled~(\%) denotes non-empty rows and Unique counts distinct non-empty strings. \emph{Oral check} corresponds to the \emph{Oral exam} aspect of the official metric.}
\label{tab:fields}
\centering
\small
\begin{tabular}{lrrl}
\toprule
Field & Filled (\%) & Unique & Observed structure \\
\midrule
Handle                  & 96 & 145 & Near visit-specific \\
Oral check              & 93 & 146 & Near visit-specific \\
Doctor advices          & 80 &  59 & Strongly templated \\
Diagnosis               & 76 &  99 & Diverse, recurrent groups \\
Treatment plan          & 59 &  88 & Diverse, incomplete \\
Main appeal             & 34 &  50 & Sparse \\
Present medical history & 28 &  43 & Sparse \\
\bottomrule
\end{tabular}
\end{table}

% =====================================================================
\section{Method}
\label{sec:method}

\subsection{Problem Formulation}

We distinguish evidence availability from evidence authority. Let $x_i$ denote the CBCT volume for case $i$ and $y_i=\{y_i^{(f)}\}_{f=1}^{7}$ its structured record. The system has access to a training-derived text prior $p=\{p^{(f)}\}$ and image-conditioned retrieved text $r_i=\{r_i^{(f)}\}$. An admission function $A_f(p,r_i,x_i)\in\{0,1\}$ determines whether the retrieved evidence may modify field $f$:
\begin{equation}
\hat{y}_i^{(f)}=
\begin{cases}
U\bigl(p^{(f)},r_i^{(f)}\bigr), & A_f(p,r_i,x_i)=1,\\
p^{(f)}, & \text{otherwise},
\end{cases}
\label{eq:admission}
\end{equation}
where $U$ is deterministic. Full authority replaces the record, absent authority retains $p$, and bounded authority restricts admission by field and content.

\subsection{Text Prior and Image Retrieval}

For each field, we select its BLEU-medoid as a stable text prior and no-image control. Retrieval then supplies optional evidence; in the submitted ECCR system, only Diagnosis may be updated. We preprocess every volume with percentile clipping, min--max normalization, and trilinear resizing to $128\times256\times256$. A frozen Med3DVLM VisionTower \cite{xin2025med3dvlm} produces a feature tensor that is mean-pooled and $\ell_2$-normalized. We standardize query and bank features with the bank mean and standard deviation before normalizing again. For query embedding $z_i$ and bank embedding $z_j$, top-1 cosine retrieval selects
\begin{equation}
j^*=\arg\max_j \frac{z_i^\top z_j}{\lVert z_i\rVert_2\lVert z_j\rVert_2},
\label{eq:retrieval}
\end{equation}
and returns $r_i=y_{j^*}$. The encoder is frozen and the 50 bank features are cached, so the pipeline requires no task-specific training.

\begin{figure}[t]
\centering
\includegraphics[width=\linewidth]{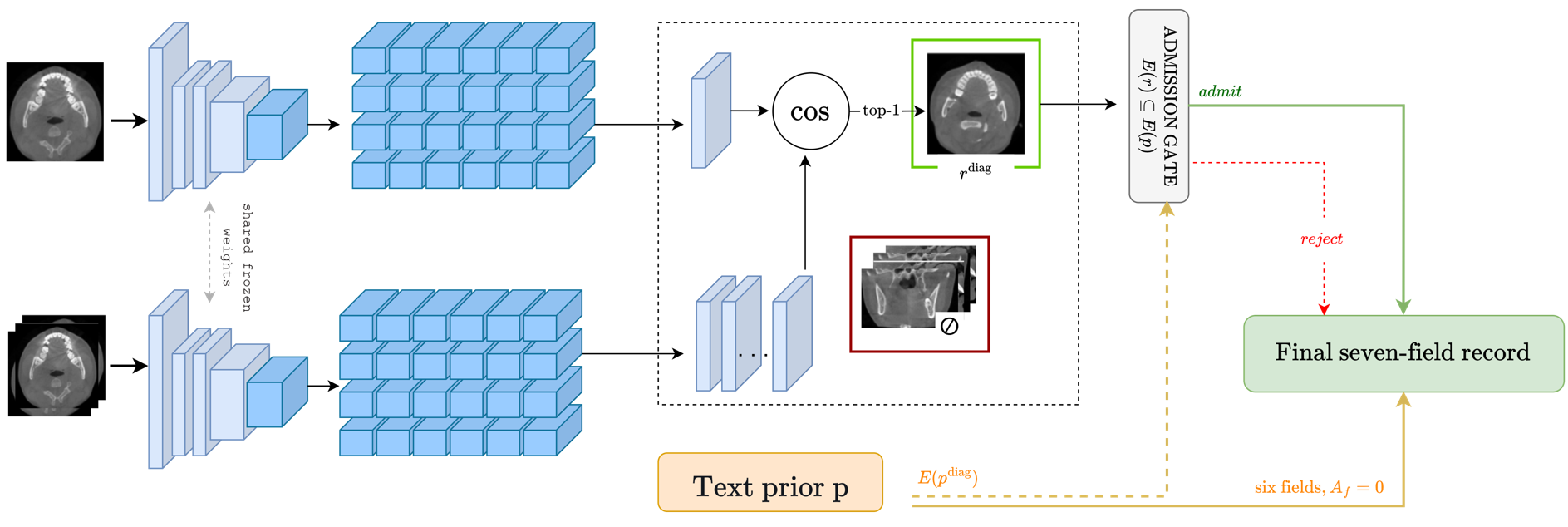}
\caption{\textbf{Retrieved evidence is controlled rather than copied.} A frozen
3D encoder retrieves a neighbour, while prior $p$ supplies all seven fields.
Retrieval may propose a \emph{Diagnosis} update only ($A_f=0$ otherwise), and
admission requires $\entity(r)\subseteq\entity(p)$; rejection retains the prior.
With $k=1$, entity consistency is the operative gate.}
\label{fig:pipeline}
\end{figure}

\subsection{Entity-Constrained Diagnosis Update}
\label{sec:bounded}

As illustrated in \Cref{fig:pipeline}, image evidence proposes rather than replaces a Diagnosis. For every non-Diagnosis field, $A_f=0$. If the candidate is admitted, we append it to the prior after exact-duplicate removal; otherwise the prior is retained. With $k=1$, neighbour agreement is degenerate, so Diagnosis-only authority and entity consistency are the operative controls.

\paragraph{Entity extraction and admission.}
Let $\entity(s)$ extract FDI positions 11--48 and dental ICD-10 codes matching \texttt{K}\emph{dd}\texttt{.}\emph{ddd} from a Diagnosis string $s$; the guarantee is deliberately limited to these measured entity types and establishes no recall property for free-text clinical concepts. The candidate is eligible only when
\begin{equation}
\entity\!\left(r_i^{(\mathrm{diag})}\right)\subseteq\entity\!\left(p^{(\mathrm{diag})}\right),
\label{eq:entity}
\end{equation}
so an admitted candidate may repeat or refine an asserted entity but cannot add a new one.

\paragraph{Entity-set invariant.}
Let $d=p^{(\mathrm{diag})}$ and $c=r_i^{(\mathrm{diag})}$. The update returns either $d$ or $d\oplus c$, where $\oplus$ denotes concatenation with a separator. Since $\entity(d\oplus c)=\entity(d)\cup\entity(c)$, \Cref{eq:entity} gives
\begin{equation}
\entity\!\left(\hat{y}_i^{(\mathrm{diag})}\right)=\entity(d).
\label{eq:invariant}
\end{equation}
Retrieval therefore cannot alter the detected entity set. The prior bounds entity precision, while its coverage caps entity recall.

% =====================================================================
\section{Experiments}
\label{sec:experiments}

\subsection{Protocol and Implementation}
\label{sec:protocol}

Our primary endpoint is the public-validation weighted score computed by Codabench against hidden references \cite{codabench}; because the split is fixed, it has no variance estimate. We additionally use a 50-case leave-one-out (LOO) proxy averaging four smoothed BLEU scores, ROUGE-L F1, tooth F1, and diagnosis-code F1. The proxy omits the official corpus-level diversity factor and is used only to compare update rules. The environment is given in \Cref{tab:env}.

\begin{table}[t]
\caption{System configuration. The proposed pipeline is parameter-free and requires only frozen encoder inference over the cached bank and query volumes.}
\label{tab:env}
\centering
\small
\setlength{\tabcolsep}{6pt}
\renewcommand{\arraystretch}{1.15}
\begin{tabularx}{\linewidth}{@{}lX@{}}
\toprule
\textbf{Component} & \textbf{Configuration} \\
\midrule
GPU & 1$\times$ NVIDIA GeForce RTX 4090 (24\,GB) \\
Encoder & Med3DVLM VisionTower \cite{xin2025med3dvlm} (frozen) \\
Optimization & No training; parameter-free inference \\
\bottomrule
\end{tabularx}
\end{table}

\paragraph{Retrieval spaces and encoder choice.}
Full-record transfer uses the joint image--text space of M3D-CLIP \cite{bai2024m3d} to match complete reports, whereas bounded admission retrieves candidate Diagnosis text from pooled Med3DVLM features \cite{xin2025med3dvlm} and relies on the text prior elsewhere. Comparing ECCR against the text-only prior therefore isolates the effect of image-conditioned evidence, since both share identical non-Diagnosis fields.

\paragraph{Official scoring.}
Four aspects are scored on a 100-point scale with weights $0.45$ (Diagnosis), $0.20$ (Oral exam), $0.20$ (Treatment plan), $0.15$ (Record completeness). A submission reproducing the ground truth exactly attains $97.4$ rather than $100$, so all totals should be read against that ceiling. The weighted content sum is then multiplied by a corpus-level template-diversity factor
\begin{equation}
\phi = 0.5 + 0.5\,d,
\label{eq:phi}
\end{equation}
where $d$ is the fraction of distinct Diagnosis texts across the 50 cases; identical long texts repeated across cases additionally have their tooth-level content excluded. A single repeated template thus incurs a double penalty, which shapes the analysis below.

\subsection{Evidence-Authority Comparison and Update Trace}
\label{sec:authority}

\Cref{tab:authority} tests the paper's central design choice. Replacing the full record with an M3D-CLIP neighbour gives 0.2237, 0.0678 below the text-only prior. Allowing only entity-compatible Diagnosis evidence raises the score by 0.0219 to 0.3134, second on the public leaderboard. The result supports selective evidence use; it does not, because of the encoder difference above, isolate authority as the sole cause of the full-record gap.

\begin{table}[t]
\caption{Public-validation comparison of evidence authority (0--1 scale). Full authority replaces the complete record, no authority emits the text prior, and ECCR permits only an entity-compatible Diagnosis update.}
\label{tab:authority}
\centering
\small
\begin{tabular}{p{3.5cm}p{4.8cm}r}
\toprule
System & Retrieval update & Score \\
\midrule
M3D-CLIP retrieval & Full: replaces the complete record & 0.2237 \\
BLEU-medoid & None: text prior only & 0.2915 \\
Ours & Diagnosis only; entity-consistent & \textbf{0.3134} \\
\bottomrule
\end{tabular}
\end{table}

\paragraph{Update-rule ablation.}
\Cref{tab:offline} examines why the final update appends rather than replaces, and reports both what the lexical proxy measures and what it cannot. Replacement is clearly harmful, costing $0.0289$. Naive and entity-safe appending differ by $0.0002$---below the resolution of a proxy that contains no term penalising an unsupported entity. The rightmost column supplies that missing term: across the 50 LOO folds the retrieved candidate carries an FDI position or ICD code absent from the prior in 40 cases. Both replacement and naive appending propagate those entities into the record, whereas \Cref{eq:invariant} guarantees zero injections. Since the official Diagnosis aspect is scored by entity-level F1, each injected entity is a false positive lowering precision: the two appending rules are equivalent on the proxy and not equivalent on the metric that decides the challenge.

\paragraph{Admission trace.}
\Cref{tab:admission} traces the guard on the 49 public-validation queries: the constraint intervenes in $63.3\%$ of retrievals, blocking 31 candidates and admitting 18 ($36.7\%$). The two traces are computed on different sets---50 LOO folds over the labeled split and 49 validation queries---and are reported separately throughout.

\begin{table}[t]
\centering
\setlength{\tabcolsep}{3pt}
\begin{minipage}[t]{0.545\linewidth}
\centering
\caption{Update-rule ablation on the strict LOO proxy (50 labeled cases, 0--1
scale, no corpus-level term). \emph{Injections} counts folds whose emitted
Diagnosis contains an FDI or ICD entity absent from the prior.}
\label{tab:offline}
\small
\begin{tabularx}{\linewidth}{Xrr}
\toprule
Method & Proxy & Inject. \\
\midrule
Text prior & 0.1831 & 0/50 \\
Diagnosis replace & 0.1542 & 40/50 \\
Naive append & \textbf{0.1833} & 40/50 \\
Entity-safe append (ours) & 0.1831 & \textbf{0/50} \\
\bottomrule
\end{tabularx}
\end{minipage}
\hfill
\begin{minipage}[t]{0.435\linewidth}
\centering
\caption{Admission trace over the 49 public-validation cases. Conflict denotes
an FDI or ICD entity absent from the prior. The gate row is degenerate at $k=1$.}
\label{tab:admission}
\small
\begin{tabularx}{\linewidth}{>{\raggedright\arraybackslash}Xrr}
\toprule
Stage / conflict & Cases & \% \\
\midrule
Retrieved evidence & 49 & 100.0 \\
Gate passed & 49 & 100.0 \\
Blocked: FDI only & 6 & 12.2 \\
Blocked: ICD only & 16 & 32.7 \\
Blocked: both & 9 & 18.4 \\
Final admitted & 18 & 36.7 \\
\bottomrule
\end{tabularx}
\end{minipage}
\end{table}

\subsection{Reconciling the Offline Proxy with the Official Score}
\label{sec:reconcile}

Smoothed BLEU and ROUGE-L reward $n$-gram overlap with a single reference and impose no penalty for an unsupported entity, so appending a short compatible clause is close to metric-neutral by construction. The parity in \Cref{tab:offline} is therefore consistent with local updates preserving baseline lexical quality; it is not evidence of the safety property, which the proxy cannot measure and which \Cref{eq:invariant} instead establishes by construction.

The official validation gain ($+0.0219$) is largely attributable to \Cref{eq:phi}. The text-only prior emits one Diagnosis template for all cases ($d=1/50$, $\phi=0.51$); admitting entity-safe evidence expands this into six distinct phrasing patterns ($d=0.12$, $\phi=0.56$). Rescaling the prior by $0.56/0.51$ gives $0.2915\times1.098\approx0.320$, which brackets the observed $0.3134$. We therefore do not claim the gain reflects improved clinical content; it reflects a legitimate reduction in template repetition obtained without altering the asserted entity set. The same mechanism exposes a structural limit of a static prior: because repeated baseline text is excluded from tooth-level scoring, a medoid prior caps tooth-level Diagnosis F1 at $0.016$ (\Cref{tab:subscores}), the single largest source of forfeited Diagnosis credit.

\subsection{Final Leaderboard Result and Score Decomposition}
\label{sec:decomposition}

Under strict penalties for entity mismatch and template repetition, scores are heavily compressed: a ground-truth submission would score $97.4$, yet the top entry reaches $17.70$ (\Cref{tab:leaderboard}). ECCR obtains $11.37$, ranking second, ahead of the third-placed system
despite a lower Diagnosis sub-score because Record completeness
contributes $3.76$ against $1.68$.

\begin{table}[t]
\caption{Final Task~3 leaderboard under the 100-point protocol (attainable maximum $97.4$). ECCR ranks second.}
\label{tab:leaderboard}
\centering
\small
\begin{tabular}{llrrrrr}
\toprule
Rank & Team & Diagnosis & Oral exam & Treatment & Completeness & Total \\
\midrule
1 & abc-abc          & 5.56 & 1.49 & 4.71 & 5.95 & 17.70 \\
2 & \textbf{Ours}    & 4.89 & 0.00 & 2.72 & 3.76 & \textbf{11.37} \\
3 & sjtu426lab-task3 & 5.29 & 0.00 & 3.12 & 1.68 & 10.08 \\
4 & jlshen           & 4.50 & 0.00 & 2.07 & 3.50 & 10.06 \\
\midrule
\multicolumn{2}{l}{Aspect weight} & 0.45 & 0.20 & 0.20 & 0.15 & 1.00 \\
\bottomrule
\end{tabular}
\end{table}

\begin{table}[t]
\caption{Sub-metric decomposition of ECCR's final score. Points are on the
100-point scale before the diversity factor; Credited applies $\phi=0.56$.
Completeness sub-metrics combine a completion rate and ROUGE-L and are reported
at aspect level only.}
\label{tab:subscores}
\centering
\small
\begin{tabular}{llrrr}
\toprule
Aspect (Points / Credited) & Sub-metric & $n$ & F1 & Points \\
\midrule
\multirow{4}{*}{Diagnosis \quad 8.73 / \textbf{4.89}}
 & Condition detection    & 47 & 0.2345 & 4.69 \\
 & Tooth--condition match & 32 & 0.0159 & 0.16 \\
 & ICD codes              & 40 & 0.1630 & 1.63 \\
 & Whole-mouth            & 20 & 0.4500 & 2.25 \\
\midrule
\multirow{3}{*}{Oral exam \quad 0.00 / \textbf{0.00}}
 & Findings detection     & 34 & 0.0000 & 0.00 \\
 & Tooth-level findings   & 21 & 0.0000 & 0.00 \\
 & Whole-mouth findings   &  7 & 0.0000 & 0.00 \\
\midrule
\multirow{3}{*}{Treatment \quad 4.85 / \textbf{2.72}}
 & Action detection       & 42 & 0.4762 & 4.76 \\
 & Tooth-level actions    & 32 & 0.0000 & 0.00 \\
 & Handled actions        & 29 & 0.0172 & 0.09 \\
\midrule
\multicolumn{5}{l}{Completeness \quad 6.72 / \textbf{3.76} \quad
(main appeal 3.18, present 1.91, past 0.00, advices 1.64)} \\
\midrule
\multicolumn{4}{l}{\textbf{Total}} & 20.30 $\rightarrow$ \textbf{11.37} \\
\bottomrule
\end{tabular}
\end{table}

\Cref{tab:subscores} makes the failure modes explicit rather than implicit. Diagnosis carries the submission at $4.89$ points, but almost all of it comes from condition-level and whole-mouth detection: tooth--condition matching contributes $0.16$ of a possible $10$, the direct consequence of the medoid cap in \Cref{sec:reconcile}. The Oral exam aspect scores zero, forfeiting $20\%$ of the available weight. This is a genuine cost of restricting $A_f$ to Diagnosis, not a design benefit: \Cref{tab:fields} shows Oral check is near visit-specific (146 unique values in 161 visits), so a corpus medoid transfers almost nothing, and we did not extend the admission rule to it because no entity guard was defined for oral findings. Finally, Past history scores zero on both completion and ROUGE-L, indicating the aggregation step leaves this field empty---an implementation gap worth $1.5$ points that a non-empty default would recover. Two of the three largest losses are thus addressable without changing the framework.

\subsection{Admission Behaviour}
\label{sec:behaviour}

\begin{figure}[t]
\centering
\includegraphics[width=0.7\linewidth]{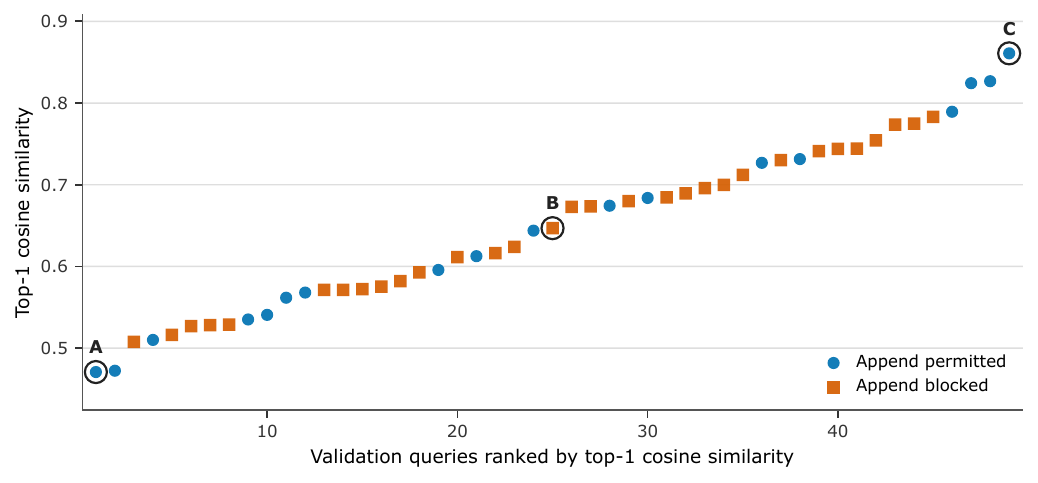}
\caption{\textbf{Similarity alone does not determine safe admission.} Validation queries are ordered by top-1 cosine similarity. Blue candidates satisfy the entity condition; orange candidates contain an FDI or ICD conflict. Blocked candidates occur across the similarity range, showing why an entity-aware rule cannot be replaced by a single similarity threshold.}
\label{fig:guard}
\end{figure}

\Cref{fig:guard} relates retrieval similarity to candidate safety. Conflicting candidates occur throughout the similarity range, so a cosine threshold cannot substitute for entity-aware verification. Retrieval capability and evidence authority address distinct failure modes: the proposed constraint prevents field coupling and unsupported entity injection, while spatial localization and prior coverage remain representation constraints for future work.

% =====================================================================
\section{Conclusion and Limitations}
\label{sec:conclusion}

ECCR combines a text prior with CBCT retrieval under governed update authority. Bounding candidate Diagnosis updates with entity preservation enriches clinical detail while preventing unverified information injection, giving a parameter-free framework for low-resource clinical record completion that requires no task-specific training.

Four limitations should be stated plainly. (i) The guarantee of \Cref{eq:invariant} covers only FDI positions and dental ICD-10 codes, and establishes no property for free-text clinical concepts. (ii) The invariant is one-sided: it bounds entity precision by the prior but caps entity recall by the prior's coverage, which is why tooth-level Diagnosis F1 remains at $0.016$. (iii) Restricting authority to Diagnosis forfeits the entire Oral exam aspect ($20\%$ of the weight); a per-field guard for oral findings is the most valuable extension we can identify. (iv) The comparison against full-record retrieval confounds authority with encoder choice (M3D-CLIP versus Med3DVLM), and the fixed single split admits no variance estimate. Future work will address region-aware volumetric representations, per-field admission rules with field-specific entity vocabularies, and learned confidence thresholds. Code will be released publicly upon acceptance.

% =====================================================================
\newpage
\bibliographystyle{splncs04}
\bibliography{ref}

@article{bai2024m3d,
  author  = {Fan Bai and Yuxin Du and Tiejun Huang and Max Q.-H. Meng and Bo Zhao},
  title   = {{M3D}: Advancing 3D Medical Image Analysis with Multi-Modal Large Language Models},
  journal = {arXiv preprint arXiv:2404.00578},
  year    = {2024}
}

@article{xin2025med3dvlm,
  author  = {Yu Xin and Gorkem Can Ates and Kuang Gong and Wei Shao},
  title   = {{Med3DVLM}: An Efficient Vision-Language Model for 3D Medical Image Analysis},
  journal = {arXiv preprint arXiv:2503.20047},
  year    = {2025}
}

@inproceedings{lewis2020rag,
  author    = {Patrick Lewis and Ethan Perez and Aleksandra Piktus and Fabio Petroni and Vladimir Karpukhin and Naman Goyal and Heinrich K{\"u}ttler and Mike Lewis and Wen-tau Yih and Tim Rockt{\"a}schel and Sebastian Riedel and Douwe Kiela},
  title     = {Retrieval-Augmented Generation for Knowledge-Intensive {NLP} Tasks},
  booktitle = {Advances in Neural Information Processing Systems},
  volume    = {33},
  pages     = {9459--9474},
  year      = {2020}
}

@article{codabench,
    title = {Codabench: Flexible, easy-to-use, and reproducible meta-benchmark platform},
    author = {Zhen Xu and Sergio Escalera and Adrien Pavão and Magali Richard and Wei-Wei Tu and Quanming Yao and Huan Zhao and Isabelle Guyon},
    journal = {Patterns},
    volume = {3},
    number = {7},
    pages = {100543},
    year = {2022}
}

@inproceedings{hamamci2024ct2rep,
  author    = {Hamamci, Ibrahim Ethem and Er, Sezgin and Menze, Bjoern},
  title     = {CT2Rep: Automated Radiology Report Generation for 3D Medical Imaging},
  booktitle = {Medical Image Computing and Computer Assisted Intervention -- MICCAI 2024},
  pages     = {476--486},
  year      = {2024},
  publisher = {Springer Nature Switzerland},
  doi       = {10.1007/978-3-031-72390-2_45}
}

@inproceedings{hosseini2025samf,
  author    = {Hosseini, Abdullah and Ibrahim, Ahmed and Serag, Ahmed},
  title     = {From Slices to Volumes: Multi-Scale Fusion of 2D and 3D Features for CT Scan Report Generation},
  booktitle = {Medical Image Computing and Computer Assisted Intervention -- MICCAI 2025},
  year      = {2025},
  publisher = {Springer Nature Switzerland},
  doi       = {10.1007/978-3-032-04978-0_26}
}

@inproceedings{miura2021factual,
  author    = {Miura, Yasuhide and Zhang, Yuhao and Tsai, Emily Bao and Langlotz, Curtis P. and Jurafsky, Dan},
  title     = {Improving Factual Completeness and Consistency of Image-to-Text Radiology Report Generation},
  booktitle = {Proceedings of the 2021 Conference of the North American Chapter of the Association for Computational Linguistics: Human Language Technologies},
  year      = {2021},
  publisher = {Association for Computational Linguistics},
  doi       = {10.18653/v1/2021.naacl-main.416},
  url       = {https://aclanthology.org/2021.naacl-main.416/}
}

@inproceedings{delbrouck2022semantic,
  author    = {Delbrouck, Jean-Benoit and Chambon, Pierre and Bl{\"u}thgen, Christian and Tsai, Emily Bao and Almusa, Omar and Langlotz, Curtis P.},
  title     = {Improving the Factual Correctness of Radiology Report Generation with Semantic Rewards},
  booktitle = {Findings of the Association for Computational Linguistics: EMNLP 2022},
  pages     = {4348--4360},
  year      = {2022},
  publisher = {Association for Computational Linguistics},
  doi       = {10.18653/v1/2022.findings-emnlp.319},
  url       = {https://aclanthology.org/2022.findings-emnlp.319/}
}

@article{jain2021radgraph,
  author  = {Jain, Saahil and Agrawal, Ashwin and Saporta, Adriel and Truong, Steven Q. H. and Duong, Du Nguyen and Bui, Tan and Chambon, Pierre and Lungren, Matthew and Ng, Andrew and Langlotz, Curtis and Rajpurkar, Pranav},
  title   = {RadGraph: Extracting Clinical Entities and Relations from Radiology Reports},
  journal = {PhysioNet},
  year    = {2021},
  doi     = {10.13026/hm87-5p47},
  url     = {https://physionet.org/content/radgraph/1.0.0/}
}

@article{zhang2020kgreport,
  author  = {Zhang, Yixiao and Wang, Xiaosong and Xu, Ziyue and Yu, Qihang and Yuille, Alan and Xu, Daguang},
  title   = {When Radiology Report Generation Meets Knowledge Graph},
  journal = {arXiv preprint arXiv:2002.08277},
  year    = {2020},
  url     = {https://arxiv.org/abs/2002.08277}
}

% =====================================================================
% CHECKLIST -- DELETE IN CAMERA-READY.
% =====================================================================
\clearpage
\begin{table}[t]
\caption{Checklist Table. Please fill out this checklist table in the answer
column. (Delete this Table in the camera-ready submission)}
\label{tab:checklist}
\centering
\small
\begin{tabularx}{\linewidth}{Xl}
\toprule
Requirements & Answer \\
\midrule
A meaningful title & Yes \\
The number of authors ($\leq$6) & 4 \\
Author affiliations and ORCID & Yes \\
Corresponding author email is presented & Yes \\
Validation scores are presented in the abstract & Yes \\
Introduction includes at least three parts: background, related work, and
motivation & Yes \\
A pipeline/network figure is provided & \Cref{fig:pipeline} \\
Pre-processing & \Cref{sec:method} \\
Strategies to data augmentation & Not used (no training) \\
Post-processing & \Cref{sec:bounded} \\
Environment setting table is provided & \Cref{tab:env} \\
Training protocol table is provided & N/A (parameter-free, no training) \\
Ablation study & \Cref{tab:authority}, \Cref{tab:offline} \\
Visualized segmentation example is provided & N/A (generation task) \\
Limitation and future work are presented & \Cref{sec:conclusion} \\
Reference format is consistent & Yes \\
Main text $\leq$8 pages (excluding references and checklist) & Yes \\
\bottomrule
\end{tabularx}
\end{table}

\end{document}